\documentclass{article}
\usepackage{spconf,graphicx}
\usepackage{amsmath,amssymb,amsfonts,bm,amsthm,balance}
\usepackage{algorithmic,algorithm}
\usepackage{textcomp}
\usepackage{xcolor}
\usepackage{subcaption}
\usepackage{dsfont}
\usepackage{cite,cleveref}
\usepackage{url}

\usepackage{color}

\newcommand{\col}[1]{\text{col}\big\{#1\big\}}
\newcommand{\grad}[1]{\nabla_{#1^\tran} }
\newcommand{\w}{{\scriptstyle\mathcal{W}}}
\newcommand{\we}{\widetilde{{\scriptstyle\mathcal{W}}}}
\newcommand{\eqdef}{\:\overset{\Delta}{=}\:}    
\DeclareMathOperator*{\argmin}{argmin}
\newtheorem{theorem}{Theorem}
\newcommand{\tran}{{\sf T}}
\title{Network Classifiers with Output Smoothing\vspace{-0.2cm}}
\name{Elsa Rizk, Roula Nassif\thanks{E. Rizk and A. H. Sayed are with the School of Engineering, EPFL, Switzerland. This work was performed while R. Nassif was a post-doctoral scholar at EPFL. She is now with the American University of Beirut, Lebanon.}, Ali H. Sayed\vspace{-0.3cm}}
\address{\small Ecole Polytechnique Federale de Lausanne, Switzerland}
\begin{document}
\ninept
\maketitle
\begin{abstract}
This work introduces two strategies for training network classifiers with heterogeneous agents. One strategy promotes global smoothing over the graph and a second strategy promotes local smoothing over neighbourhoods. It is assumed that the feature sizes can vary from one agent to another, with some agents observing insufficient attributes to be able to make reliable decisions on their own. As a result, cooperation with neighbours is necessary. However, due to the fact that the feature dimensions are different across the agents, their classifier dimensions will also be different. This means that cooperation cannot rely on combining the classifier parameters. We instead propose smoothing the outputs of the classifiers, which are the predicted labels. By doing so, the dynamics that describes the evolution of the network classifier becomes more challenging than usual because the classifier parameters end up appearing as part of the regularization term as well. We illustrate performance by means of computer simulations. 
\end{abstract}
\vspace{-0.1cm}
\vspace{-0.25cm}
\section{Introduction and Related Work}
\label{sec:intro}
\vspace{-0.2cm}
This work studies the problem of classification by a collection of networked agents. It is assumed that the network is heterogeneous in that the feature sizes can vary from one agent to another, with some agents observing insufficient attributes to make reliable decisions on their own. As a result, cooperation with neighbours is necessary. However, due to the fact that the feature dimensions are different across the agents, their classifier dimensions will also be different. This means that cooperation cannot rely on combining the classifier parameters as in~\cite{sayed2014adaptive,sayed2014adaptation,towfic2013distributed,nassif2018distributed,tuck2019distributed}. We instead focus on smoothing the outputs of the classifiers, which are the predicted labels. By doing so, the dynamics that describes the evolution of the network classifier becomes more challenging than usual because the classifier parameters now appear as part of the regularization term as well. We explain how to address these challenges in the article. 

Specifically, we formulate two network classification problems. In one case, we assume the labels vary ``smoothly'' over the graph. In other words, we assume that the number of jumps from one label to another is small. This scenario is common since, often in practice, agents in close proximity observe more or less related attributes as well as objects belonging to the same class. For this situation, we formulate an optimization problem that enforces global graph-wide smoothing across all agents. In a second case, we assume that all agents regularly observe data from the same class so that the network is synchronized and working as a classification cluster. This situation is also common in practice when sensing networks are tasked with monitoring a common environment. For this problem, we introduce an optimization problem that involves instead a form of local smoothing across neighbourhoods.

Thus, this work introduces two strategies to train a network of heterogeneous agents to perform classification. One strategy promotes global smoothing over neighbourhoods. and a second strategy promotes local smoothing over the graph. In the sequel, we explain how these formulations differ from earlier works in the literature. 

For now, it is useful to note that network classifiers can have a multitude of applications, including in weather forecasting \cite{RoulaWeather}, surveillance and object recognition \cite{zhang2016distributed}, brain activity detection \cite{cox2003functional}, and healthcare monitoring.

\vspace{-0.4cm}
\subsection{Previous Work}
\label{subsec:lit}
\vspace{-0.2cm}
There exist several good works in the literature on combining classifiers. One notable example is the class of boosting techniques, such as AdaBoost \cite{freund1999short}. However, these techniques are centralized and do not exploit any underlying graph structure. There exist other works that focus on incorporating the graph structure into the learning algorithm. These methods, from within the framework of representation learning \cite{hamilton2017representation,ahmed2013distributed,hamilton2017inductive,scarselli2008graph,Goyal_2018,Cai}, are more focused on embedding the graph structure into the feature vector.  Once this is accomplished, the data is passed into traditional learning methods such as SVMs or neural networks. Some of these methods involve node embedding, which aims at representing the location and neighbourhood of each node in the feature vector; while subgraph embedding represents entire  subgraphs as low dimensional feature vectors. 

The main issue with these methods is that we must first learn an embedding of the graph to generate feature vectors, before we can  learning algorithms on these vectors. In our approach, we avoid the graph embeding step and exploit the graph structure directly by devising different learning algorithms. We formulate distributed optimization problems that enforce regularization on the graph and forces the agents to collaborate as dictated by the topology that binds them. We comment on further distinctions from prior work at the end of the next section after we formulate the network classification problems. 

\vspace{-0.4cm}
\subsection{Contribution}
\label{subsec:contribution}
\vspace{-0.2cm}
Consider an undirected graph $\mathcal{G}$ that consists of $K$ nodes. Let $A \in \mathbb{R}^{K\times K}$ denote the graph combination matrix, which is assumed to be doubly stochastic ($\mathds{1}^\tran A = \mathds{1}$, $A\mathds{1} = \mathds{1}^\tran$). Each node $k$ has access to a  set of $N$ data points consisting of labels $\gamma_k(n) \in \{\pm 1\}$ and feature vectors $h_{k,n} \in \mathbb{R}^{M_k}$. The attributes within the feature vectors can be different across the nodes. The size of the feature vectors can also differ across the graph, which motivates using a subscript $k$ in $M_k$. Also, the labels can vary across the nodes, with some clusters in the network observing data arising from class $+1$ and other clusters observing data arising from class $-1$. Since the feature sizes at the agents are different, some agents may not have sufficient information for reliable classification on their own, which motivates the need for cooperation. One issue is to decide on how cooperation should be carried out since the classifiers will have different dimensions and cannot be aggregated directly by the agents. We clarify these questions in the sequel.

Let ${\cal L}$ denote the graph Laplacian matrix defined by ${\cal L} = D - A$, where $D$ is a diagonal matrix with elements
\begin{equation}
	[D]_{kk} = \sum_{\ell\in \mathcal{N}_k} a_{k\ell},
	\vspace{-0.2cm}
\end{equation}
with $\mathcal{N}_k$ denoting the neighbourhood of agent $k$ and $a_{k\ell}\geq 0$ denoting the weight scaling the data moving from agent $k$ to agent $\ell$. 

Let $w_k\in\mathbb{R}^{M_k}$ denote the classification vector at agent $k$. Let also $\widehat{\gamma}_k(n)$ denote the estimate for the label at agent $k$ and sample $n$. We  start from the following aggregate optimization problem:
\vspace{-0.2cm}
\begin{equation}\label{eq:P1}
	\w^o = \argmin_{\w} J(\w) \eqdef \frac{1}{N}\sum_{n=0}^{N-1}\sum_{k=1}^K Q(\gamma_k(n),\widehat{\gamma}_k(n)) +  \eta R(w_k),
\end{equation}
where we have collected all classifiers into  $\w = \col{w_k}_{k=1}^K$, and the loss function $Q(\cdot)$ is assumed to be differentiable, $\nu-$strongly convex, and with $\delta-$Lipschitz gradients. The term $R(\cdot)$ represents a regularization factor, with $\eta>0$. We assume in this article that $R(\cdot) $ is differentiable for simplicity of presentation. Non-smooth regularizes can also be used with minimal adjustments to the algorithms by using easier subgradients or incorporating  proximal steps. 

Although nodes may be observing data belonging to different class labels, we shall assume smooth transitions across the graph. It is natural in applications to assume that such predictions vary slowly across adjacent nodes. Specifically, if we let $\widehat{\gamma}_k(n) = h_{k,n}^\tran w_k$ denote the prediction for the label at agent $k$ at time $n$, we shall enforce that these predictions vary smoothly over the graph by modifying \eqref{eq:P1} to the following form: 

\vspace{-0.2cm}
\begin{equation}\label{eq:P2}
\begin{split}
		\w^{\star}_{\rho} = \argmin_{\w} J^s(\w) \eqdef  \frac{1}{N}&\sum_{n=0}^{N-1}\sum_{k=1}^K \Big[ Q(\gamma_k(n),\widehat{\gamma}_k(n)) \\& + \eta R(w_k)\Big] + \frac{1}{2}\rho\widehat{\gamma}_n^T {\cal L} \widehat{\gamma}_n, 
\end{split}
\end{equation}
where $\widehat{\gamma}_n = \text{col}\{\widehat{\gamma}_k(n)\}_{k=1}^K$ and $\rho>0$ is a regularization factor. This form includes an additional regularization term in terms of the Laplacian matrix and exploits the fact that the Laplacian helps enforce smoothness among connected agents~\cite{anis2018sampling,shuman2013emerging,smola2003kernels}. We will also consider a second problem formulation where we can continue with the optimization problem \eqref{eq:P1} but define instead the prediction for $\widehat{\gamma}_k(n)$  \eqref{eq:avg} by averaging the predictions from the neighbours having the same label as $k$, where ${\cal C}_k = \{\ell \in {\cal N}_k | \gamma_{\ell}=\gamma_k\}$.

\begin{equation}\label{eq:avg}
	\widehat{\gamma}_k(n) = \frac{1}{|{\cal C}_k|} \sum_{\ell \in {\cal C}_k}a_{\ell k} h_{\ell,n}^\tran w_{\ell}
\end{equation}
The main difference between formulations \eqref{eq:P2} and \eqref{eq:P1}-\eqref{eq:avg} is that \eqref{eq:P2} imposes smoothness on a global graph-wide scale, while \eqref{eq:P1}-\eqref{eq:avg} imposes smoothness on a local neighbourhood scale. 

There are two aspects that set our formulations apart from prior works in the literature. First, the feature vectors have different dimensions across the agents with some agents having insufficient information to operate alone. Different from the works in \cite{bYing,heinze2018preserving} where the network corresponds to one classifier and supervised learning is performed under a distributed features assumption, in the current work we assume that each agent corresponds to one classifier and that the outputs of the classifiers are smooth over the graph. As a result, the local classifiers also have different dimensions and they cannot be combined together directly to strengthen performance. We are therefore dealing with a heterogeneous network situation. This is one reason why we perform smoothing on labels and not on the parameter space. The second issue is that this formulation  leads to a more challenging dynamic equation describing the evolution of the network, which in turn makes the performance analysis more demanding. This is because the unknown classifiers also appear as part of the regularization term.

\section{Distributed Classification Algorithms}
\label{sec:Sol} 

\begin{figure}[H]
\begin{subfigure}{.25\textwidth}
  \centering
  \includegraphics[width=\linewidth]{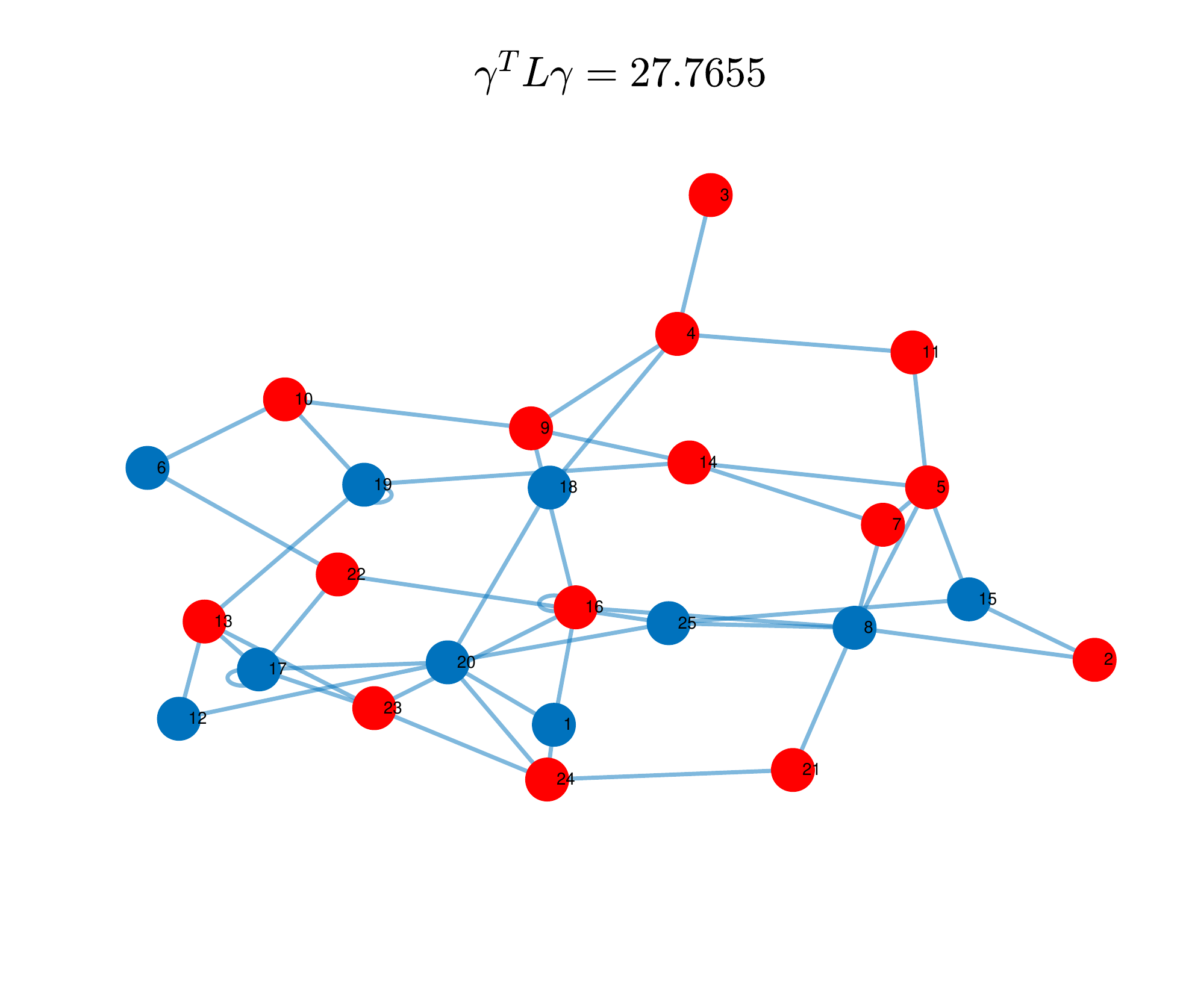}
  \label{fig:nonsmoothEG}
\end{subfigure}%
\begin{subfigure}{.25\textwidth}
  \centering
  \includegraphics[width=\linewidth]{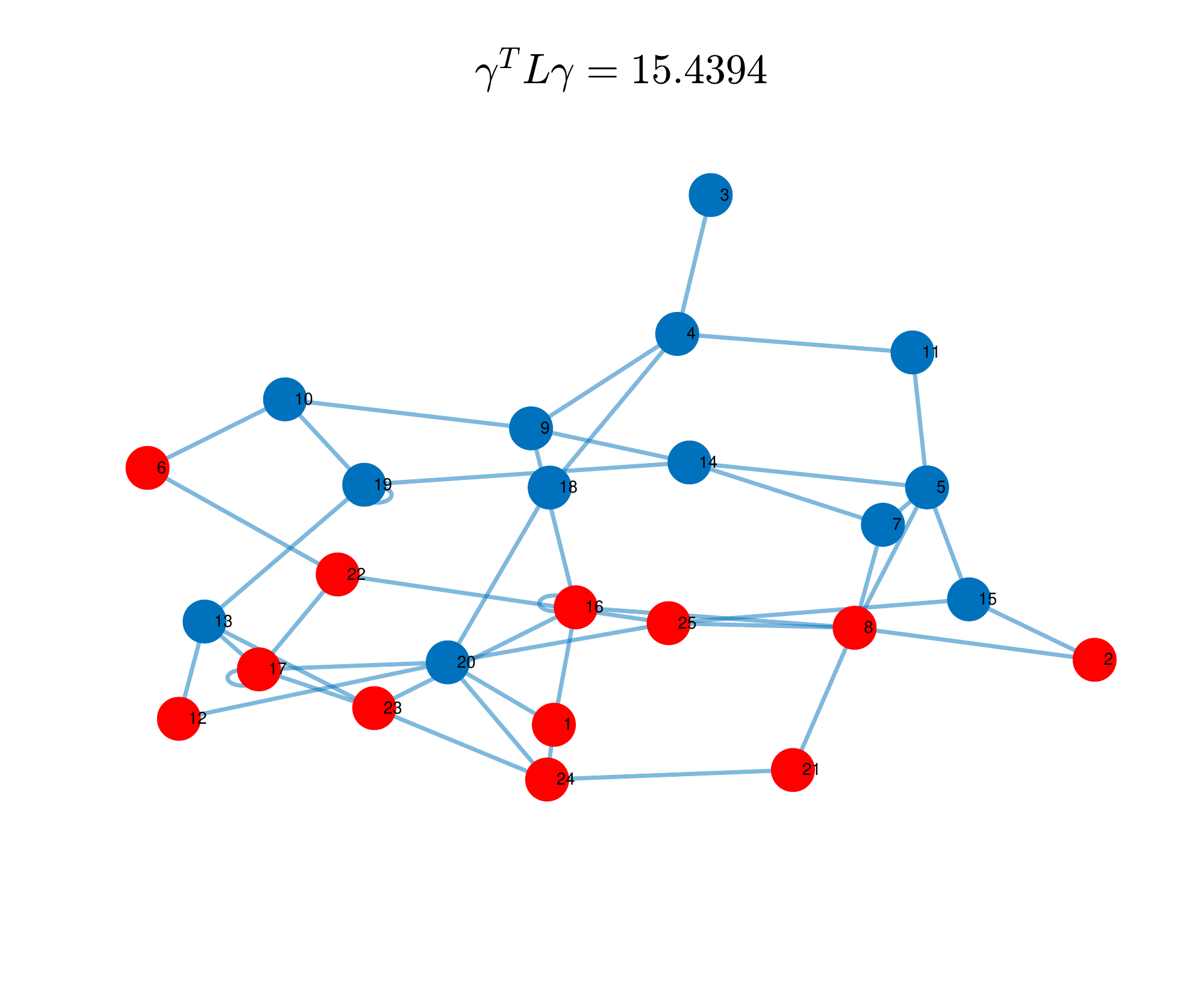}
  \label{fig:smoothEG}
\end{subfigure}
\vspace{-1cm}
\caption{Two graphs with different smoothness levels. }\label{fig:SmoothnessEG}
\end{figure}

Let us consider first the regularized problem \eqref{eq:P2}, which enforces global smoothing across the graph. For this setting, we assume the labels $\gamma_k$ vary smoothly over the graph, meaning that  the number of jumps that occur from a node with label $+1$ to a node with label $-1$ is small. One example to this effect is illustrated in Fig. \ref{fig:SmoothnessEG}, where the network on the left has multiple jumps, while the network on the right has fewer jumps. Note that the value of the regularization term $\gamma^\tran {\cal L}\gamma$ is smaller in the second case.  Examining the graph on the right, we see that some agents observe one label (say, the red agents), while another group of agents observes the second label (say, the blue agents). In this work, we shall assume that these clusters continue to act as a group continually, i.e., agents in them observe the same labels continually. For example, assume the graph is classifying ``cars'' and ``airplanes.'' It is assumed that the red cluster, as a group, observes either cars or airplanes together, and similarly for the blue cluster. This situation maintains smoothness. Situations where  class observations vary arbitrarily across all agents lead to non-smooth scenarios and are less frequent; we do not cover this scenario here. It is more natural that local clusters observe data arising from the same label, which is in line with the assumption of smoothness. 

Writing down a distributed stochastic gradient algorithm for minimizing \eqref{eq:P2} leads to the listing of Algorithm \ref{alg:smoothness}. In this implementation, a random index $n$ is selected at each iteration and the gradient vector of the cost function is evaluated in an incremental manner, by first computing a stochastic-gradient approximation for the loss term, which is denoted by $g_{k,\bm{n}}$. This gradient is further updated by adding the correction that arises from differentiating the regularizer and smoothing terms in the cost \eqref{eq:P2}. This results in $g'_{k,\bm{n}}$, which is then used to update $\bm{w}_{k,i-1}$ to $\bm{w}_{k,i}$. Observe that coordination among the agents occurs during the evaluation of the second gradient correction. Once the network is trained, then during testing each agent estimates its label by averaging the label information from neighbours using \eqref{eq:avg}. This is necessary because even after training, some agents will still be unable to classify on their own because they do not have sufficient features. 

\begin{algorithm}
\begin{algorithmic}
\caption{(Network classifier with graph smoothing)}\label{alg:smoothness}
\STATE{\textbf{given $N$ data pairs} $\{\gamma_k(n),h_{k,n}\}_{n=0}^{N-1}$ \textbf{for every} $k=1,2,\cdots,K.$ \\
\textbf{initialize} $\bm{w}_{k,-1}$ \textbf{for every} $k = 1,2,\cdots,K.$\;}
\FOR{each iteration $i=1,2,\cdots$}\STATE{ \vspace{-0.3cm}
\FOR{each agent $k=1,2,\cdots,K$} \STATE {$\bm{n}$ = random index of data to be used at iteration $i$ \\
$\widehat{\gamma}_k(\bm{n}) = h_{k,\bm{n}}^\tran \bm{w}_{k,i-1}$ \\
$g_{k,\bm{n}} = \nabla Q(\gamma_k(\bm{n}),\widehat{\gamma}_k(\bm{n}))+ \eta \nabla R(\bm{w}_{k,i-1})$ \\
$g_{k,\bm{n}}' = g_{k,n} - \mu \rho \sum\limits_{\ell\in\mathcal{N}_k}a_{k\ell}\big(\widehat{\gamma}_k(\bm{n}) - \widehat{\gamma}_{\ell}(\bm{n}) \big)h_{k,\bm{n}}	$
\\ $\bm{w}_{k,i} = \bm{w}_{k,i-1} - \mu g'_{k,\bm{n}}$} \ENDFOR 	
}\ENDFOR
\end{algorithmic}
\end{algorithm}

We can repeat the same derivation for problem \eqref{eq:P1}-\eqref{eq:avg}, leading to Algorithm \ref{alg:avg}. In this implementation, each agent estimates its label by using the local averaging structure \eqref{eq:avg}. The vector $g_{k,\bm{n}}$ refers to a stochastic approximation for the loss term, and is used to update $\bm{w}_{k,i-1}$ to $\bm{w}_{k,i}$. Again, once the classifiers are trained, and during testing, the agents employ \eqref{eq:avg} to estimate their labels. This second implementation is more suitable for scenarios where the entire network acts as a cluster with all agents observing features from the same class. We will illustrate these effects in the simulation section. 

\begin{algorithm}[H]
\begin{algorithmic}
\caption{(Network classifier with local smoothing)}\label{alg:avg}
\STATE{\textbf{given $N$ data pairs} $\{\gamma_k(n),h_{k,n}\}_{n=0}^{N-1}$ \textbf{for every} $k=1,2,\cdots,K.$ \\
\textbf{initialize} $\bm{w}_{k,-1}$ \textbf{for every} $k = 1,2,\cdots,K.$\;}
\FOR{each iteration $i=1,2,\cdots$}\STATE{ \vspace{-0.3cm}
\FOR{each agent $k=1,2,\cdots,K$} \STATE {$\bm{n}$ = random index of data to be used at iteration $i$ \\
$\widehat{\gamma}_k(\bm{n}) = \frac{1}{|{\cal C}_k|}\sum_{\ell \in {\cal C}_k}a_{\ell k}h_{\ell,\bm{n}}^\tran \bm{w}_{\ell,i-1}$ \\
$g_{k,\bm{n}} = \nabla Q(\gamma_k(\bm{n}),\widehat{\gamma}_k(\bm{n}))+ \eta \nabla R(\bm{w}_{k,i-1})$ \\
$\bm{w}_{k,i} = \bm{w}_{k,i-1} - \mu g_{k,\bm{n}}$} \ENDFOR 	
}\ENDFOR
\end{algorithmic}
\end{algorithm}

\section{Network Evolution}
\vspace{-0.2cm}
We now examine the evolution of the network and show that the algorithms are able to converge to a small neighbourhood around their optimal solutions. Due to space limitations, we focus on the first algorithm since the analysis for the second one is similar. 

We collect the estimates from across the network into the column vector $\bm{\w}_i = \col{\bm{w}_{k,i}}_{k=1}^K$, and define ${\cal H}_i =$ blckdiag$\{h_{1,\bm{n}}^\tran,$ $ h_{2,\bm{n}}^\tran,\cdots, h_{K,\bm{n}}^\tran\}$, where $\bm{n}$ is a function of $i$. Then, the recursion for Algorithm \ref{alg:smoothness} can be written as:

\begin{equation}\label{eq:P2Rec}
	\bm{\w}_i 
	= (I-\mu\rho \mathcal{H}_i {\cal L}\mathcal{H}_i^\tran)\bm{\w}_{i-1} - \mu\widehat{\grad{\mbox{\tiny $\w$}} J}(\bm{\w}_{i-1}),
	\vspace{-0.3cm}
\end{equation} 
where 
\begin{equation}
\begin{split}
	\widehat{\grad{\mbox{\tiny $\w$}} J}(\bm{\w}_{i-1}) = \col{&\nabla Q(\gamma_k(n),\widehat{\bm{\gamma}}_k(n)) \\ &+\eta\nabla R(\bm{w}_{k,i-1})}_{k=1}^K.\end{split}
\end{equation}
The stochastic gradient noise process is given by 

\begin{equation}
\begin{split}
	\bm{s}_{i}(\w) &= \widehat{\grad{\mbox{\tiny $\w$}}J^s}(\w) - \grad{\mbox{\tiny $\w$}}J^s(\w) \\
	&= \widehat{\grad{\mbox{\tiny $\w$}}J}(\w) + \rho {\cal H}_i{\cal L}{\cal H}_i^\tran \w \\
	&\quad - \grad{\mbox{\tiny $\w$}}J(\w) - \rho \frac{1}{N}\sum_{m=0}^{N-1}{\cal H}_m{\cal L }{\cal H}_m^\tran \w.
\end{split}
\end{equation} 
It is customary in the literature on distributed optimization to assume that the gradient noise satisfies the conditions \cite{sayed2014adaptation}:
\begin{subequations}
\begin{align}
	\mathbb{E}[\bm{s}_i (\bm{\w}_{i-1})| \bm{\w}_{i-1}] &= 0 \label{eq:cond1}\\
	\mathbb{E}[\Vert \bm{s}_i (\bm{\w}_{i-1}) \Vert^2 | \bm{\w}_{i-1}] & \leq \beta_s^2 \Vert \bm{\we}_{i-1}\Vert^2 + \sigma_s^2 \label{eq:cond2},
\end{align}
\end{subequations}
where $\bm{\we}_i = \w^{\star}_{\rho}-\bm{\w}_i$,
for some positive constants $\beta_s^2,\sigma_s^2$. These conditions are automatically satisfied for important cases of interest. 
\vspace{-0.3cm}
\begin{theorem}
	Consider the stochastic recursion \eqref{eq:P2Rec}. For step size values  satisfying $\mu < \frac{2\nu}{\delta^2 + \beta_s^2}$ (i.e., for $\mu$ small enough),
it holds that $\mathbb{E}\Vert \bm{\we}_i\Vert^2$ converges exponentially fast according to the recursion:
\begin{equation}\label{eq:thrm1Rec}
	\mathbb{E}\Vert \bm{\we}_i\Vert^2 \leq \lambda\mathbb{E}\Vert\bm{\we}_{i-1}\Vert^2 + \mu^2 \sigma_s^2,
\end{equation}
where $ \lambda = 1-2\nu\mu + (\delta^2 + \beta_s^2)\in [0,1)$. It follows from \eqref{eq:thrm1Rec} that, for sufficiently small step sizes:
\begin{align}
	\limsup_{n\to\infty} \mathbb{E}\Vert \bm{\we}_i\Vert^2 &\leq O(\lambda^i) + O(\mu) \\
	\limsup_{n\to\infty} \mathbb{E} J^s(\bm{\w}_i) -  J^s(\w^{\star}_{\rho}) &\leq O(\lambda^i) + O(\mu)
\end{align} 
with the convergence of $\mathbb{E} J^s(\bm{\w}_i)$ towards the $O(\mu)-$ neighbourhood around $J^s(\w^{\star}_{\rho})$ occurring at the exponential rate $\lambda^i$ as well.
\qed
\end{theorem}

\noindent
{\em Sketch of proof}. 
	Using the already established result for a single agent under uniform sampling found in \cite[p.~345]{sayed2014adaptation}, what we need to show is that the aggregate algorithm \eqref{eq:P2Rec} satisfies the same conditions in that result.
	First, by writing 
	\begin{align}
	\overline{\mathcal{B}} &= \frac{1}{N}\sum_{m=0}^{N-1} \mathcal{B}_m = \frac{1}{N}\sum_{m=0}^{N-1}{\mathcal H}_m{\cal L}{\mathcal H}_m^\tran   \\
	p_i &= \text{col}\Bigg\{\frac{1}{N}\sum_{m=0}^{N-1}\nabla Q(\bm{w}_{k,i-1},\gamma_k(m))\Bigg\}_{k=1}^K \\
	q^{\star} &= \col{\nabla Q(w_{\rho,k}^{\star})}_{k=1}^K \\
	\end{align} 
	we get the error recursion: 
\begin{equation}\label{eq:errRec}
	\bm{\we}_i = (I-\mu\rho \overline{\mathcal{B}})\bm{\we}_{i-1} + \mu p_{i-1} + \mu \bm{s}_i(\bm{\w}_{i-1})+\mu\rho \overline{\mathcal{B}}\w^{\star}_{\rho}.
	\vspace{-0.2cm}
\end{equation}
The conditions on the gradient noise hold since it can be verified that
\begin{align}
	\mathbb{E}[\bm{s}_i(\bm{\w}_{i-1}) | \bm{\w}_{i-1}] &= 0 \\
	\mathbb{E}\big[\Vert \bm{s}_i(\bm{\w}_{i-1})\Vert^2 | \bm{\w}_{i-1}\big]  &\leq \beta_s^2 \Vert \bm{\we}_{i-1}\Vert^2 + \sigma_s^2 ,
\end{align} 
where $\beta_s^2 = 8\delta^2 + 4\rho^2\mathbb{E}\Vert {\cal B}_i-\overline{{\cal B}}\Vert_F^2$ and $\sigma_s^2 = 4\mathbb{E}\Vert q^{\star} - \rho {\cal B}_i\w^{\star}_{\rho}\Vert^2$. 
\qed

\section{Experimental Results}
\label{sec:Results}

For the first experiment, we simulate the following scenario. Consider a network of sensors observing the same phenomenon, such as tracking an object that could be either a car or a plane. Since the sensors are focusing on different characteristics, then their feature vectors are of different sizes arising from the same label. The individual agents are not able to classify on their own if they have insufficient features. 

\begin{figure}[H]
 \begin{center}
 	\includegraphics[scale=0.18]{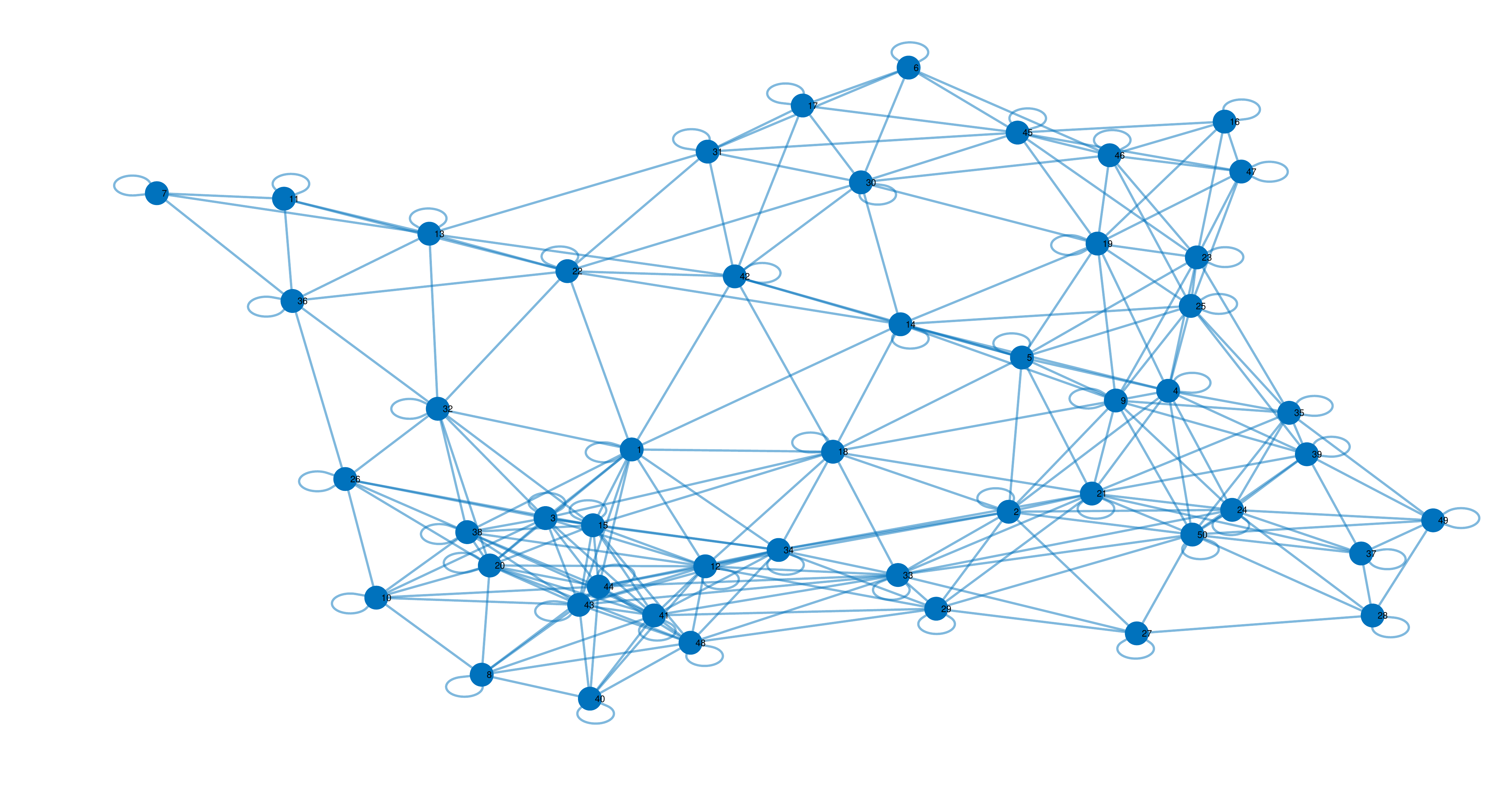}
 	\vspace{-0.3cm}
		\caption{Network structure with uniform labels used in the simulations.}\label{fig:graph}
 	\end{center}
 \end{figure}

Thus, consider an undirected graph consisting of $K=50$ agents. The graph is generated by randomly placing the nodes in a coordinate system, and then taking the neighbourhood of radius $0.3$ for each node. The combination matrix $A$ is generated using the Metropolis rule. The resulting network is shown in Fig. \ref{fig:graph}.
\vspace{-0.5cm}
We assume the feature vectors have sizes up to $M=5$ attributes.  Each agent receives $N=200$ samples of data $\{\gamma_k(n),$ $h_{k,n}\}_{n=0}^{199}$ with $h_{k,n} \in \mathbb{R}^{M_k}$ and $\gamma_k(n)\in \{\pm 1\}$. The data is generated as follow: For each agent $k$, the size of the feature vector $M_k$ is generated randomly from a discrete uniform distribution on $\{1,2,3,4,5\}$. Second, each attribute $m \in\{1,2,3,4,5\}$ is a Gaussian random variable with mean $\mu_m$ and variance $\sigma_m^2$. Then, the $N$ feature vectors of each agent are generated from joint Gaussian distributions of the features. As for the labels, they are generated randomly by a fair coin toss; it is assumed that for each sample $n=0,1,\cdots,N-1$, all agents have the same label.

We consider the logistic risk as our cost function with $\ell 2$-norm regularization; thus, equation \eqref{eq:P1} becomes~\cite{hosmer2013applied,theodoridis2008pattern}: 
\begin{equation}\label{eq:logisticRisk}
		\min_{\w} \frac{1}{N} \sum_{n=0}^{N-1} \sum_{k=1}^K \Big[ \log(1+e^{-\gamma_k(n)\widehat{\gamma}_k(n)}) + \eta \Vert w_k\Vert^2 \Big].
\end{equation} 

\begin{figure}[H]
\begin{subfigure}[b]{0.25\textwidth}
\centering
  \includegraphics[width=\linewidth ]{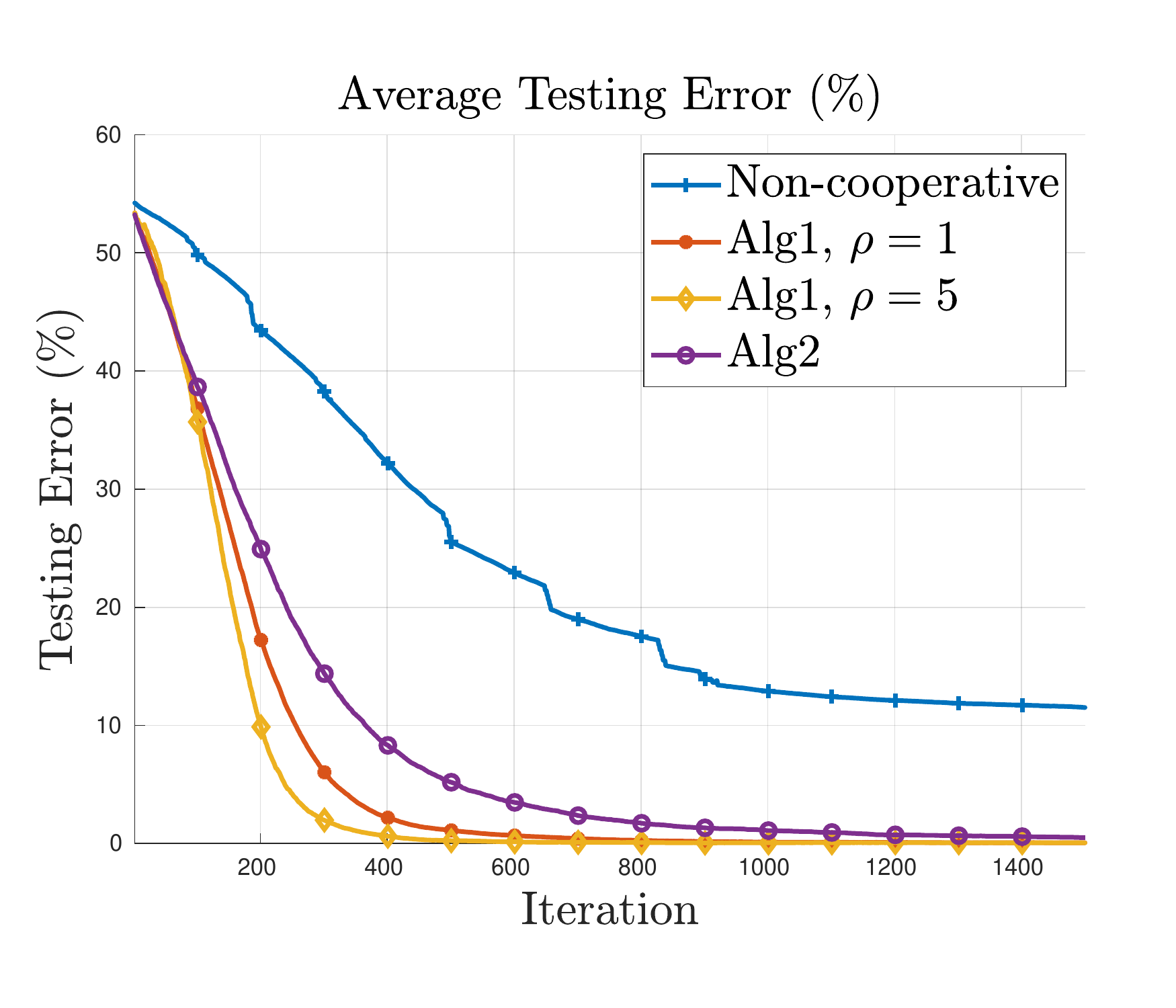}
  \caption{Uniform labels across the\\ network.}
  \label{fig:TE_uni}
  \end{subfigure}%
  \begin{subfigure}[b]{0.25\textwidth}
 	\centering
 	\includegraphics[width=\linewidth]{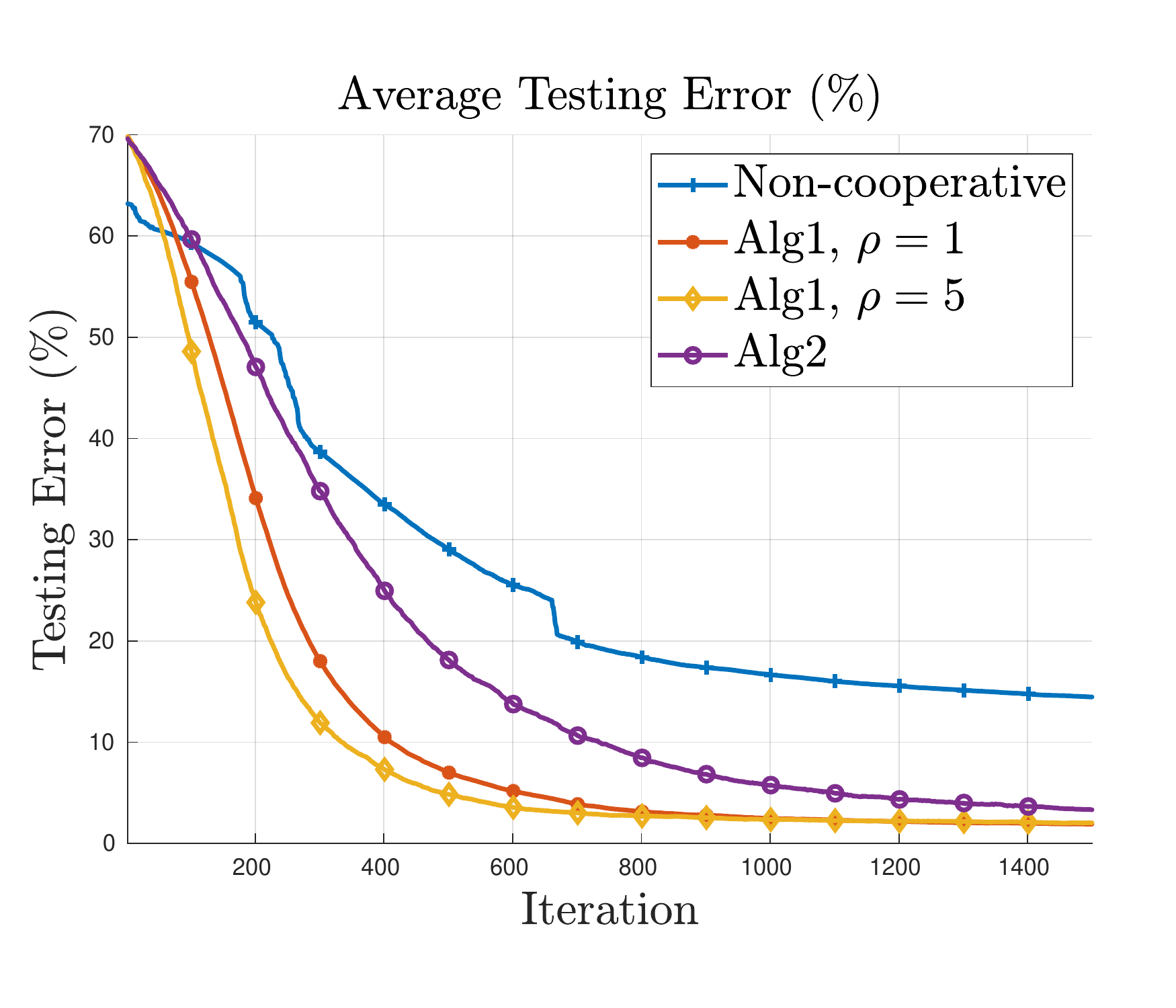}
 	\caption{Non-uniform labels across \\the network.}
 	\label{fig:TE_mult}
  \end{subfigure}
\vspace{-0.7cm}
  \caption{ Average testing errors on simulated data.}\label{fig:TE_uniL}
\end{figure}

The two algorithms are run by first splitting the data into a training set consisting of $70\%$ of the data and a testing set consisting of $30\%$ of the data. A step size of $\mu = 5\times 10^{-3}$ is used, with $\eta = 0.1$. We perform 5 passes over the data. The average testing error across the nodes (Fig. \ref{fig:TE_uniL}) is plotted. Algorithm \ref{alg:smoothness} is run both with and without ($\rho=0$). From Fig. \ref{fig:TE_uni}, we see that the non-cooperative algorithm ($\rho = 0$) performs the worst. When collaboration is enforced, the network classifier performs significantly better. Comparing Algorithms \ref{alg:smoothness} and \ref{alg:avg}, we see that the former converges faster. 

\begin{figure}[H]
	\begin{center}
		\includegraphics[scale=0.18]{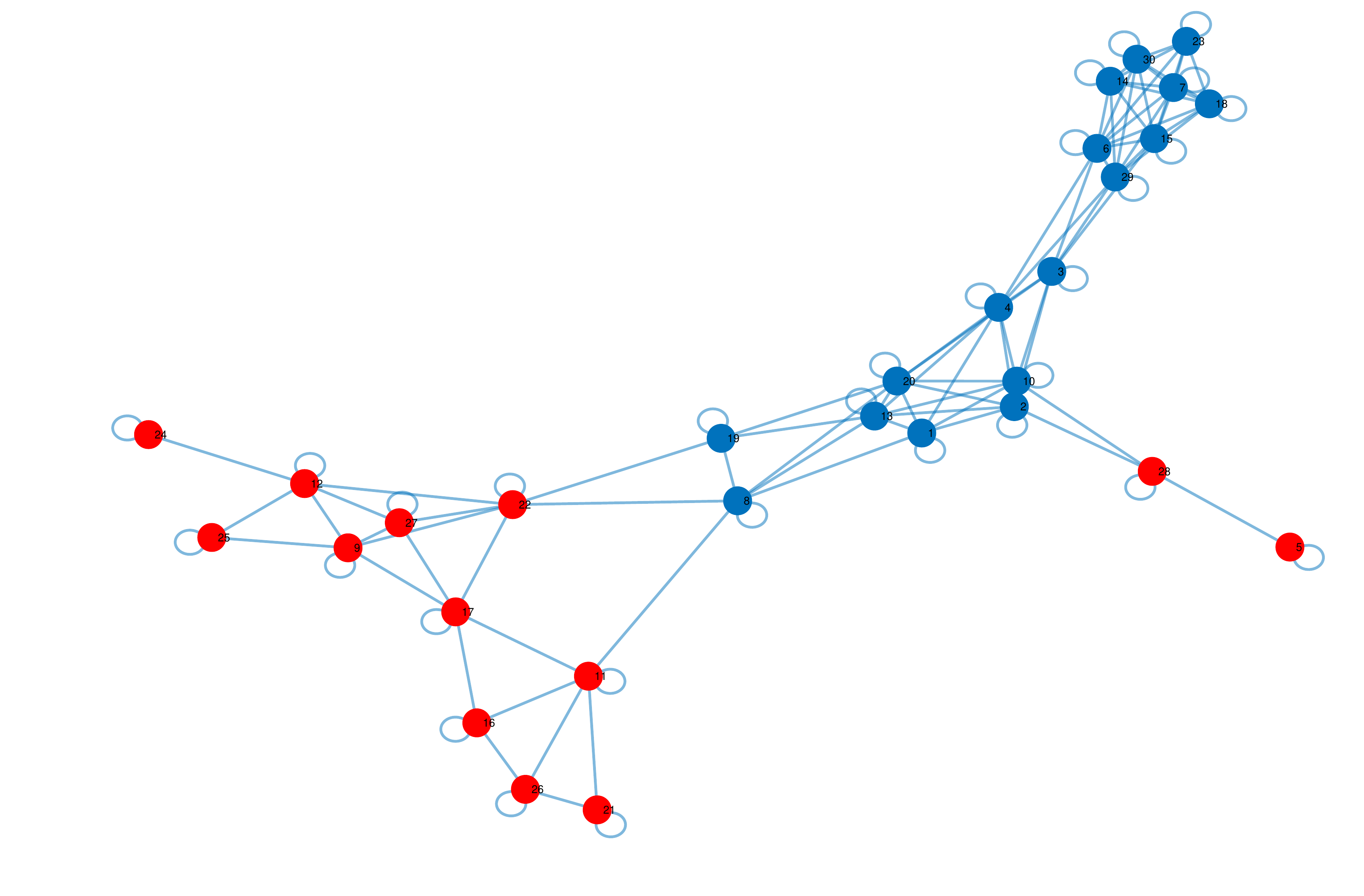}
		\vspace{-0.3cm}
		\caption{Network structure with  non-uniform labels used in the simulations.}\label{fig:graph_mult}
	\end{center}
\end{figure}

\begin{figure}[H]
	\begin{subfigure}[b]{0.25\textwidth}
		\centering
		\includegraphics[width=\linewidth ]{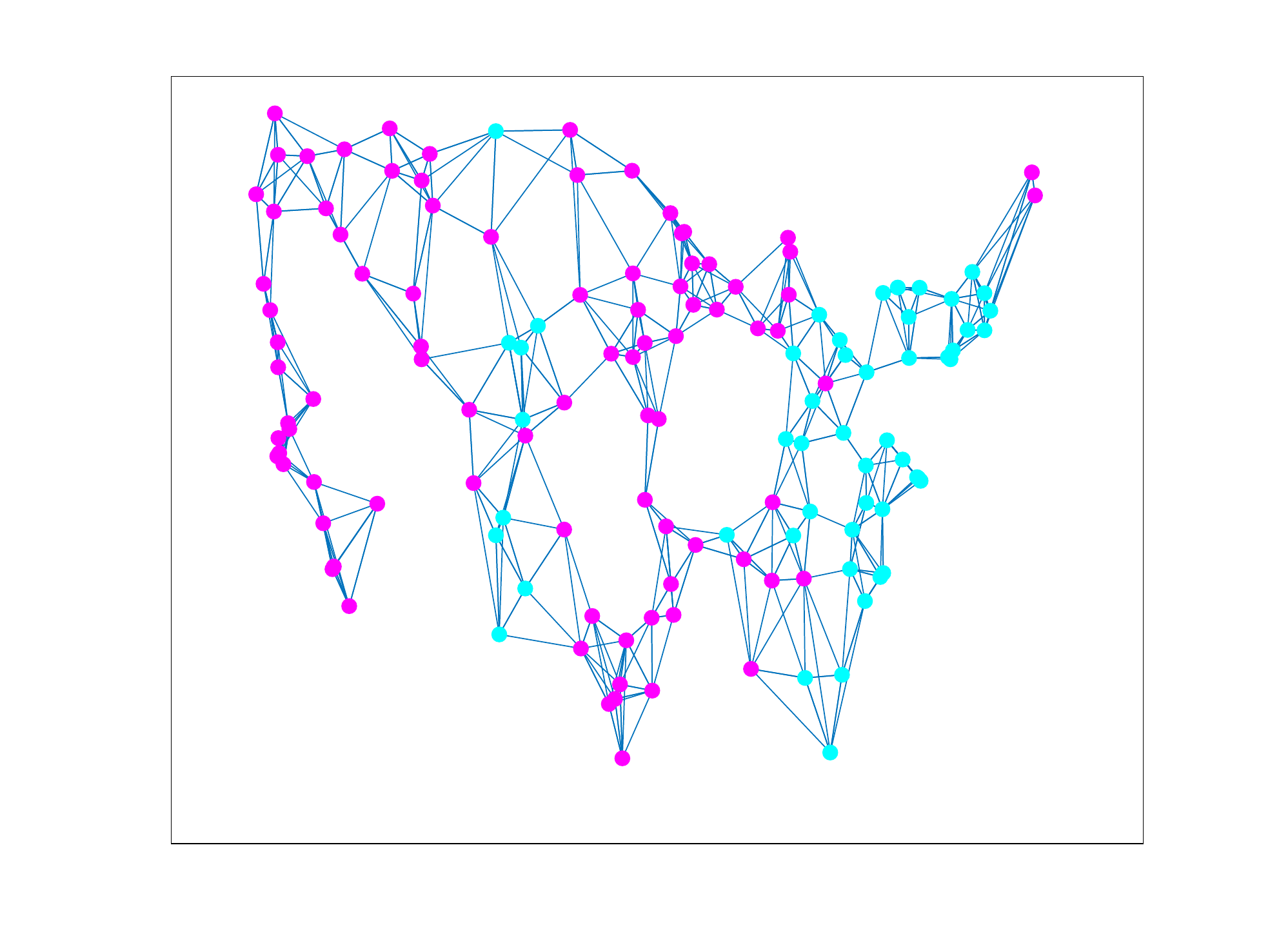}
		\caption{True labels.}\label{fig:Weath_true}
	\end{subfigure}%
	\begin{subfigure}[b]{0.25\textwidth}
		\centering 
		\includegraphics[width=\linewidth]{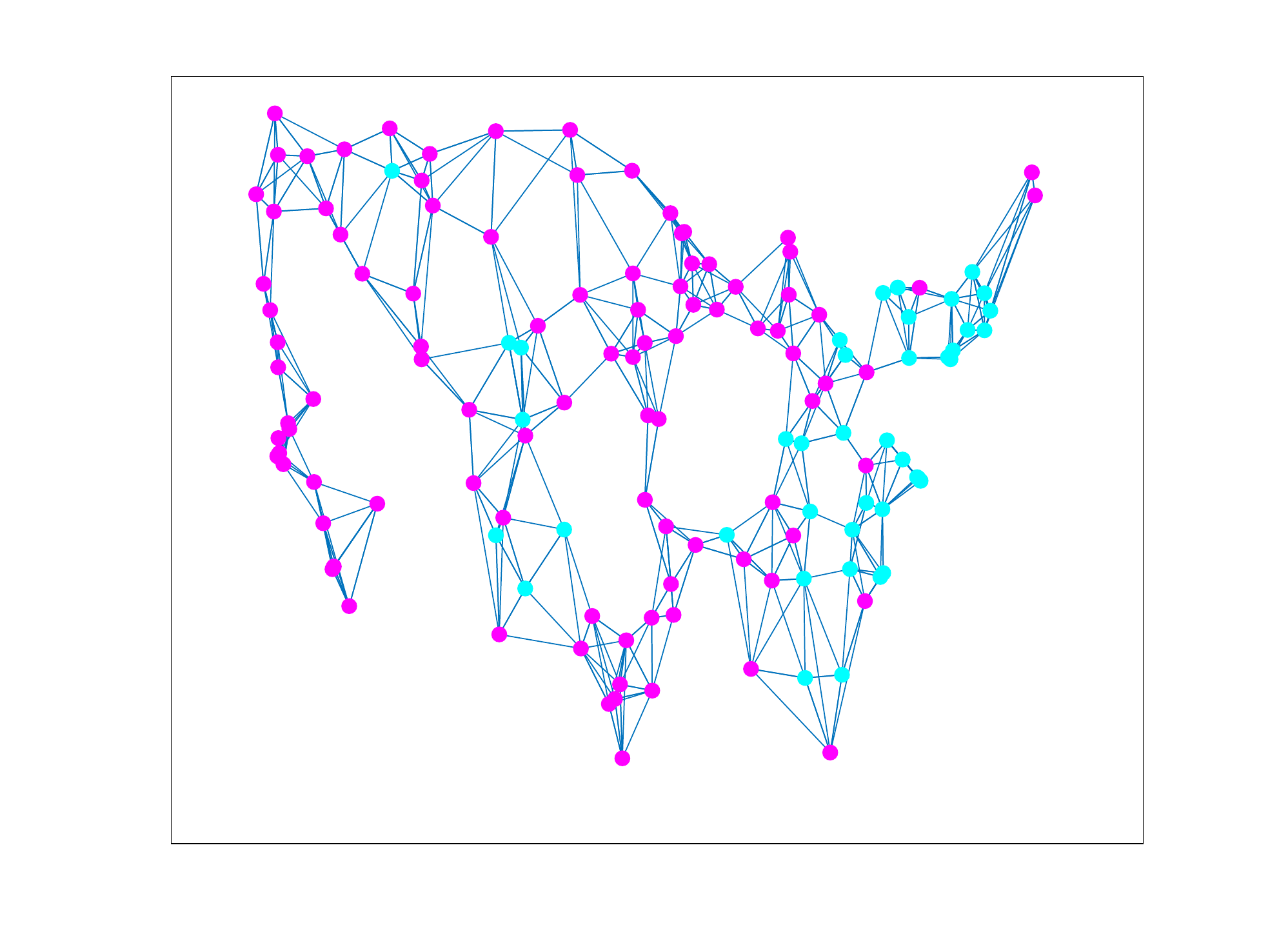}
		\caption{Predicted labels.}\label{fig:Weath_pred}
	\end{subfigure}
	\caption{Weather labels for the 1st of July, 2013: blue corresponding to a rainy day and purple to a non-rainy day.}\label{fig:weath}
\end{figure}

Next, we consider a scenario where one part of the network is observing data from one class while another part is observing data from another class, as shown in Fig. \ref{fig:graph_mult}. We observe similar results as in the previous case, except that now the cooperative algorithms do not reach zero testing errors, but they still outperform the non-cooperative algorithm. 
\vspace{-0.6cm}
Finally, we consider a real weather dataset consisting of a collection of daily recordings over 13 years, taken from $K = 139$ sensors located in different weather stations across the United States \cite{weatherData}. A 6-nearest neighbour distance graph was created to represent the network of sensors. Each sensor has a set of $N = 3288$ samples consisting of features indicating weather there is rain or not, and a feature vector of size $M = 5$. In this experiment, we only run Algorithm \ref{alg:smoothness} and compare it to the non-cooperative case, since this data does not fall under the assumptions made for the second algorithm (neighbouring agents may observe different labels over time). We set the step size to $\mu = 3\times 10^{-4}$, and the regularizer parameters $\eta = 1\times 10^{-5}$ and $\rho = 0.3$. The non-cooperative solution achieved an average testing error of 0.2001, while Algorithm \ref{alg:smoothness} achieved 0.1851. In Fig. \ref{fig:weath}, for the 1st of July, 2013, we represent the true labeled graph on the left and the predicted one on the right. Specifically, for this day, the testing error achieved was 0.1079. 
\vspace{-0.3cm}

\section{CONCLUDING REMARKS}
\label{sec:Concl}
\vspace{-0.3cm}
In this work, we proposed two algorithms that solve multi-agent classification problems while enforcing collaboration and two forms of smoothing: global over the graph and local over the neighbourhoods. We provided convergence results for one of the algorithms, while indicating that the same approach applies to the other one. We noticed that collaboration through smoothing improves performance. Future work involves examining different optimization formulations that allow for other forms of smoothing coupled with automatic clustering of agents into classes. 


%

\pagebreak




\bibliographystyle{IEEEbib}
{\balance{\bibliography{ICASP2020}}}

\begin{thebibliography}{10}

\bibitem{sayed2014adaptive}
A.~H. Sayed,
\newblock ``Adaptive networks,''
\newblock {\em Proc. of the IEEE}, vol. 102, no. 4, pp. 460--497, Apr. 2014.

\bibitem{sayed2014adaptation}
A.~H. Sayed,
\newblock ``Adaptation, learning, and optimization over networks,''
\newblock {\em Foundations and Trends in Machine Learning}, vol. 7, no. 4-5,
  pp. 311--801, 2014.

\bibitem{towfic2013distributed}
Z.~J. Towfic, J.~Chen, and A.~H. Sayed,
\newblock ``On distributed online classification in the midst of concept
  drifts,''
\newblock {\em Neurocomputing}, vol. 112, pp. 138 -- 152, 2013.

\bibitem{nassif2018distributed}
R.~Nassif, S.~Vlaski, and A.~H. Sayed,
\newblock ``Distributed inference over multitask graphs under smoothness,''
\newblock in {\em Proc. IEEE International Workshop on Signal Processing
  Advances in Wireless Communications}, Kalamata, Greece, Jun. 2018, pp. 1--5.

\bibitem{tuck2019distributed}
J.~Tuck, S.~Barratt, and S.~Boyd,
\newblock ``A distributed method for fitting {L}aplacian regularized stratified
  models,''
\newblock {\em Available as arXiv:1904.12017v2}, Jun. 2019.

\bibitem{RoulaWeather}
R.~{Nassif}, C.~{Richard}, J.~{Chen}, and A.~H. {Sayed},
\newblock ``Distributed diffusion adaptation over graph signals,''
\newblock in {\em IEEE International Conference on Acoustics, Speech and Signal
  Processing}, Calgary, Alberta, Canada, April 2018, pp. 4129--4133.

\bibitem{zhang2016distributed}
W.~Zhang, D.~Zhao, L.~Xu, Z.~Li, W.~Gong, and J.~Zhou,
\newblock ``Distributed embedded deep learning based real-time video
  processing,''
\newblock in {\em IEEE International Conference on Systems, Man, and
  Cybernetics}, Budapest, Hungary, 2016, pp. 1945--1950.

\bibitem{cox2003functional}
D.~D. Cox and R.~L. Savoy,
\newblock ``Functional magnetic resonance imaging (fmri)``brain reading'':
  Detecting and classifying distributed patterns of fmri activity in human
  visual cortex,''
\newblock {\em Neuroimage}, vol. 19, no. 2, pp. 261--270, 2003.

\bibitem{freund1999short}
Y.~Freund, R.~Schapire, and N.~Abe,
\newblock ``A short introduction to boosting,''
\newblock {\em Journal-Japanese Society For Artificial Intelligence}, vol. 14,
  no. 771-780, pp. 1612, 1999.

\bibitem{hamilton2017representation}
W.~L. Hamilton, R.~Ying, and J.~Leskovec,
\newblock ``Representation learning on graphs: Methods and applications,''
\newblock {\em IEEE Data Engineering Bulletin}, vol. 40, no. 3, pp. 52--74,
  2017.

\bibitem{ahmed2013distributed}
A.~Ahmed, N.~Shervashidze, S.~Narayanamurthy, V.~Josifovski, and A.~J. Smola,
\newblock ``Distributed large-scale natural graph factorization,''
\newblock in {\em Proc. International Conference on World Wide Web}, Rio de
  Janeiro, Brazil, 2013, ACM, pp. 37--48.

\bibitem{hamilton2017inductive}
W.~Hamilton, Z.~Ying, and J.~Leskovec,
\newblock ``Inductive representation learning on large graphs,''
\newblock in {\em Advances in Neural Information Processing Systems},
  California, USA, 2017, pp. 1024--1034.

\bibitem{scarselli2008graph}
F.~Scarselli, M.~Gori, A.~C. Tsoi, M.~Hagenbuchner, and G.~Monfardini,
\newblock ``The graph neural network model,''
\newblock {\em IEEE Transactions on Neural Networks}, vol. 20, no. 1, pp.
  61--80, 2008.

\bibitem{Goyal_2018}
P.~Goyal and E.~Ferrara,
\newblock ``Graph embedding techniques, applications, and performance: A
  survey,''
\newblock {\em Knowledge-Based Systems}, vol. 151, pp. 78–94, Jul 2018.

\bibitem{Cai}
H.~{Cai}, V.~W. {Zheng}, and K.~C. {Chang},
\newblock ``A comprehensive survey of graph embedding: Problems, techniques,
  and applications,''
\newblock {\em IEEE Transactions on Knowledge and Data Engineering}, vol. 30,
  no. 9, pp. 1616--1637, Sep. 2018.

\bibitem{anis2018sampling}
A.~Anis, A.~El Gamal, S.~Avestimehr, and A.~Ortega,
\newblock ``A sampling theory perspective of graph-based semi-supervised
  learning,''
\newblock {\em IEEE Transactions on Information Theory}, vol. 64, no. 4, pp.
  2322 -- 2342, 2018.

\bibitem{shuman2013emerging}
D.~I. Shuman, S.~K. Narang, P.~Frossard, A.~Ortega, and P.~Vandergheynst,
\newblock ``The emerging field of signal processing on graphs: {E}xtending
  high-dimensional data analysis to networks and other irregular domains,''
\newblock {\em IEEE Signal Processing Magazine}, vol. 30, no. 3, pp. 83--98,
  2013.

\bibitem{smola2003kernels}
A.~J. Smola and R.~Kondor,
\newblock ``Kernels and regularization on graphs,''
\newblock in {\em Learning Theory and Kernel Machines}, B.~Sch{\"o}lkopf and
  M.~K. Warmuth, Eds. 2003, pp. 144--158, Springer.

\bibitem{bYing}
B.~{Ying}, K.~{Yuan}, and A.~H. {Sayed},
\newblock ``Supervised learning under distributed features,''
\newblock {\em IEEE Transactions on Signal Processing}, vol. 67, no. 4, pp.
  977--992, Feb 2019.

\bibitem{heinze2018preserving}
C.~Heinze-Deml, B.~McWilliams, and N.~Meinshausen,
\newblock ``Preserving privacy between features in distributed estimation,''
\newblock {\em stat}, vol. 7, no. 1, pp. 189, 2018.

\bibitem{hosmer2013applied}
D.~W. Hosmer and S.~Lemeshow,
\newblock {\em Applied {L}ogistic {R}egression},
\newblock Wiley, NJ, 2nd edition, 2000.

\bibitem{theodoridis2008pattern}
S.~Theodoridis and K.~Koutroumbas,
\newblock {\em Pattern {R}ecognition},
\newblock Academic Press, 4th edition, 2008.

\bibitem{weatherData}
J.~H. Lawrimore, M.~J. Menne, B.~E. Gleason, C.~N. Williams, D.~B. Wuertz, and
  R.~S. Vose,
\newblock ``Global historical climatology network--monthly (ghcn-m),''
  \url{ftp:/ftp.ncdc.noaa.gov/pub/data/gsod}.

\end{thebibliography}

\end{document}